\documentclass{bmvc2k}

\usepackage[]{graphicx}
\usepackage{times}
\usepackage{epsfig}
\usepackage{amsmath}
\usepackage{amssymb}
\usepackage{tabularx}
\usepackage{array}
\usepackage{rotating}
\usepackage{color}
\usepackage{enumerate}
\usepackage{booktabs}
\def\httilde{\mbox{\tt\raisebox{-.5ex}{\symbol{126}}}}

\begin{document}

\newif\ifsupp
\suppfalse 
\ifsupp
    \title{Recognizing Image Style: Supplementary Materials}
    \runninghead{Karayev et al.}{Recognizing Image Style: Supplementary}
\else
    \title{Recognizing Image Style}
    \runninghead{Karayev et al.}{Recognizing Image Style}
\fi
\makeatletter{}\def\httilde{\mbox{\tt\raisebox{-.5ex}{\symbol{126}}}}
\addauthor{Sergey Karayev}{}{1}
\addauthor{Matthew Trentacoste}{}{2}
\addauthor{Helen Han}{}{1}
\addauthor{Aseem Agarwala}{}{2}
\addauthor{Trevor Darrell}{}{1}
\addauthor{Aaron Hertzmann}{}{2}
\addauthor{Holger Winnemoeller}{}{2}
\addinstitution{University of California, Berkeley}
\addinstitution{Adobe}
 
\maketitle

\ifsupp
    \listoffigures
    \listoftables
\else
    \begin{abstract}
    \makeatletter{}The style of an image plays a significant role in how it is viewed, but style has received little attention in computer vision research.
We describe an approach to predicting style of images, and perform a thorough evaluation of different image features for these tasks.
We find that features learned in a multi-layer network generally perform best -- even when trained with object class (not style) labels.
Our large-scale learning methods results in the best published performance on an existing dataset of aesthetic ratings and photographic style annotations.
We present two novel datasets: 80K Flickr photographs annotated with 20 curated style labels, and 85K paintings annotated with 25 style/genre labels.
Our approach shows excellent classification performance on both datasets.
We use the learned classifiers to extend traditional tag-based image search to consider stylistic constraints, and demonstrate cross-dataset understanding of style.
 
    \end{abstract}

    \makeatletter{}\section{Introduction}
\makeatletter{}\begin{figure}[t]
\small{
\centering
\begin{subfigure}[t]{0.48\linewidth}
    \begin{tabular}{cc}
        \includegraphics[width=.43\linewidth]{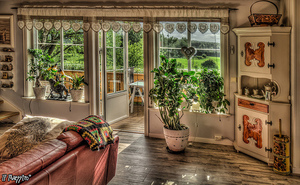} &
    \includegraphics[width=.43\linewidth]{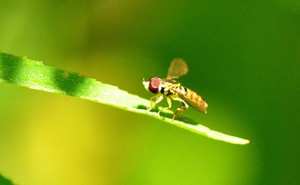} \\
    HDR & Macro \\
        \includegraphics[width=.43\linewidth]{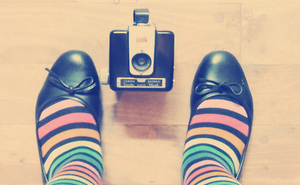} &
    \includegraphics[width=.43\linewidth]{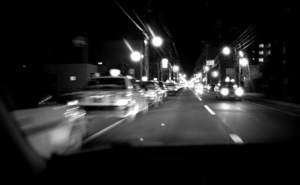} \\
    Vintage & Noir \\
        \includegraphics[width=.43\linewidth]{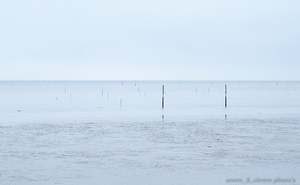} &
    \includegraphics[width=.43\linewidth]{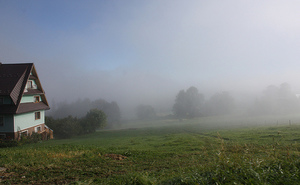} \\
    Minimal & Hazy \\
        \includegraphics[width=.43\linewidth]{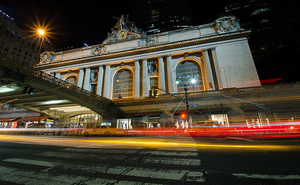} &
    \includegraphics[width=.43\linewidth]{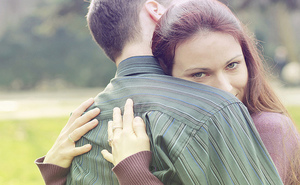} \\
    Long Exposure & Romantic \\
    \end{tabular}
    \caption{
        Flickr Style: 80K images covering 20 styles.
    }\label{fig:flickr_style_examples}
\end{subfigure}\hspace{2em}\begin{subfigure}[t]{0.48\linewidth}
    \begin{tabular}{cc}
        \includegraphics[width=.43\linewidth]{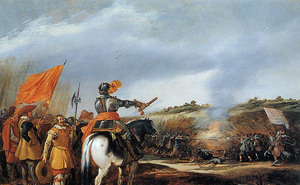} &
    \includegraphics[width=.43\linewidth]{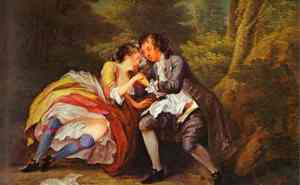} \\
    Baroque & Roccoco \\
        \includegraphics[width=.43\linewidth]{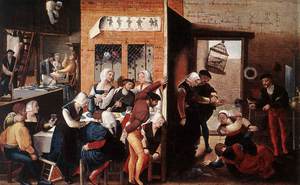} &
        \includegraphics[width=.43\linewidth]{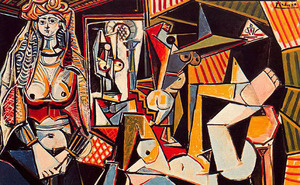} \\
    Northern Renaissance & Cubism \\
        \includegraphics[width=.43\linewidth]{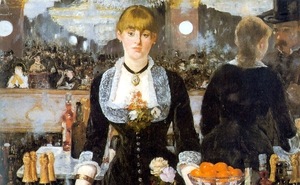} &
        \includegraphics[width=.43\linewidth]{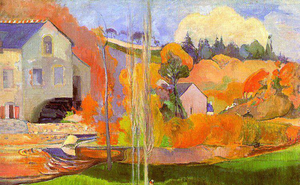} \\
    Impressionism & Post-Impressionism \\
        \includegraphics[width=.43\linewidth]{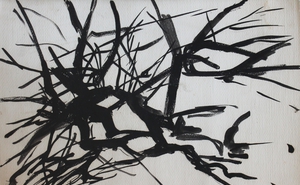} &
    \includegraphics[width=.43\linewidth]{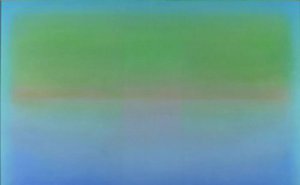} \\
    Abs. Expressionism & Color Field Painting \\
    \end{tabular}
    \caption{
        Wikipaintings: 85K images for 25 art genres.
    }\label{fig:wikipaintings_style_examples}
\end{subfigure}
}
\caption{
    Typical images in different style categories of our datasets.
}\label{fig:style_examples}
\end{figure}

Deliberately-created images convey meaning, and \textit{visual style} is often a significant component of image meaning.
For example, a political candidate portrait made in the lush colors of a Renoir painting tells a different story than if it were in the harsh, dark tones of a horror movie.
Distinct visual styles are apparent in art, cinematography, advertising, and have become extremely popular in amateur photography, with apps like Instagram leading the way.
While understanding style is crucial to image understanding, very little research in computer vision has explored visual style.

Although is it very recognizable to human observers, visual style is a difficult concept to rigorously define.
Most academic discussion of style has been in an art history context, but the distinctions between, say, Rococo versus pre-Rafaelite style are less relevant to modern photography and design.
There has been some previous research in image style, but this has principally been limited to recognizing a few, well-defined optical properties, such as depth-of-field.

We define several different \textit{types} of image style, and gather a new, large-scale dataset of photographs annotated with style labels.
This dataset embodies several different aspects of visual style, including photographic techniques  (``Macro,'' ``HDR''), composition styles (``Minimal,'' ``Geometric''), moods (``Serene,'' ``Melancholy''), genres (``Vintage,'' ``Romantic,'' ``Horror''), and types of scenes (``Hazy,'' ``Sunny'').
These styles are not mutually exclusive, and represent different attributes of style.
We  also gather a large dataset of visual art (mostly paintings) annotated with art historical style labels, ranging from Renaissance to modern art.
\autoref{fig:style_examples} shows some samples.

We test existing classification algorithms on these styles, evaluating several state-of-the-art image features.
Most previous work in aesthetic style analysis has  used hand-tuned features, such as color histograms.
We find that deep convolutional neural network (CNN) features perform best for the task.
This is surprising for several reasons: these features were trained on object class categories (ImageNet), and many styles appear to be primarily about color choices, yet the CNN features handily beat color histogram features.
This leads to one conclusion of our work: mid-level features derived from object datasets are generic for style recognition, and superior to hand-tuned features.

We compare our predictors to human observers, using Amazon Mechanical Turk experiments, and find that our classifiers predict Group membership at essentially the same level of accuracy as Turkers.
We also test on the AVA aesthetic prediction task \cite{Murray-CVPR-2012}, and show that using the ``deep'' object recognition features improves over the state-of-the-art results.

\paragraph{Applications and code.}
First, we demonstrate an example of using our method to search for images by style.
This could be useful for applications such as product search, storytelling, and creating slide presentations.
In the same vein, visual similarity search results could be filtered by visual style, making possible queries such as ``similar to this image, but more Film Noir.''
Second, style tags may provide valuable mid-level features for other image understanding tasks.
For example, there has increasing recent effort in understanding image meaning, aesthetics, interestingness, popularity, and emotion (for example, \cite{Gygli-ICCV-2013,Isola-CVPR-2011,joo2014,khosla2014}), and style is an important part of meaning.
Finally, learned predictors could be a useful component in modifying the style of an image.

All data, trained predictors, and code (including results viewing interface) are available at \url{http://sergeykarayev.com/recognizing-image-style/}.
 
    \makeatletter{}\section{Related Work}

Most research in computer vision addresses recognition and reconstruction, independent of image style.
A few previous works have focused directly on image composition, particularly on the high-level attributes of beauty, interestingness, and memorability.

Most commonly, several previous authors have described methods to predict aesthetic quality of photographs.
Datta et al.~\cite{Datta-ECCV-2006}, designed visual features to represent concepts such as colorfulness, saturation, rule-of-thirds, and depth-of-field, and evaluated aesthetic rating predictions on photographs; The same approach was further applied to a small set of Impressionist paintings~\cite{Li-SP-2009}.
The feature space was expanded with more high-level descriptive features such as ``presence of animals'' and ``opposing colors'' by Dhar et al., who also attempted to predict Flickr's proprietary ``interestingness'' measure, which is determined by social activity on the website~\cite{Dhar-CVPR-2011}.
Gygli et al.~\cite{Gygli-ICCV-2013} gathered and predicted human evaluation of image interestingness, building on work by Isola et al.~\cite{Isola-CVPR-2011}, who used various high-level features to predict human judgements of image memorability.
In a similar task, Borth et al.~\cite{Borth-MM-2013} performed sentiment analysis on images using object classifiers trained on adjective-noun pairs.

Murray et al.~\cite{Murray-CVPR-2012} introduced the Aesthetic Visual Analysis (AVA) dataset, annotated with ratings by users of DPChallenge, a photographic skill competition website.
The AVA dataset contains some photographic style labels (e.g., ``Duotones,'' ``HDR''), derived from the titles and descriptions of the photographic challenges to which photos were submitted.
Using images from this dataset, Marchesotti and Peronnin~\cite{Marchesotti-BMVC-2013} gathered bi-grams from user comments on the website, and used a simple sparse feature selection method to find ones predictive of aesthetic rating.
The attributes they found to be informative (e.g., ``lovely photo,'' ``nice detail'') are not specific to image style.

Several previous authors have developed systems to classify classic painting styles, including \cite{keren2002,shamir2010}.
These works consider only a handful of styles (less than ten apiece), with styles that are visually very distinct, e.g., Pollock vs.~Dal\'{\i}.
These datasets comprise less than 60 images per style, for both testing and training.
Mensink \cite{Mensink2014} provides a larger dataset of artworks, but does not consider style classification.
 
    \makeatletter{}\section{Data Sources}

Building an effective model of photographic style requires annotated training data.  To our knowledge, there is only one existing dataset annotated with visual style, and only a narrow range of photographic styles is represented~\cite{Murray-CVPR-2012}.
We would like to study a broader range of styles, including different \textit{types} of styles ranging from genres, compositional styles, and moods.
Morever, large datasets are desirable in order to obtain effective results, and so we would like to obtain data from online communities, such as Flickr.

\paragraph{Flickr Style.}
Although Flickr users often provide free-form tags for their uploaded images, the tags tend to be quite unreliable.
Instead, we turn to Flickr groups, which are community-curated collections of visual concepts.
For example, the Flickr Group ``Geometry Beauty'' is described, in part, as ``Circles, triangles, rectangles, symmetric objects, repeated patterns'', and contains over 167K images at time of writing; the ``Film Noir Mood'' group is described as ``Not just  black and white photography, but a dark, gritty, moody feel...'' and comprises over 7K images.

At the outset, we decided on a set of 20 visual styles, further categorized into types:
\begin{itemize}
\setlength{\itemsep}{-.5em}
\item \textbf{Optical techniques:} Macro, Bokeh, Depth-of-Field, Long Exposure, HDR
\item \textbf{Atmosphere:} Hazy, Sunny
\item \textbf{Mood:} Serene, Melancholy, Ethereal
\item \textbf{Composition styles:} Minimal, Geometric, Detailed, Texture
\item \textbf{Color:} Pastel, Bright
\item \textbf{Genre:} Noir, Vintage, Romantic, Horror
\end{itemize}

For each of these stylistic concepts, we found at least one dedicated Flickr Group with clearly defined membership rules.
From these groups, we collected 4,000 positive examples for each label, for a total of 80,000 images.
Example images are shown in \autoref{fig:flickr_style_examples}.
The exact Flickr groups used are given in \autoref{tab:flickr_groups}.

The derived labels are considered clean in the positive examples, but may be noisy in the negative examples, in the same way as the ImageNet dataset \cite{Deng-CVPR-2009}.
That is, a picture labeled as \emph{Sunny} is indeed \emph{Sunny}, but it may also be \emph{Romantic}, for which it is not labeled.
We consider this an unfortunate but acceptable reality of working with a large-scale dataset.
Following ImageNet, we still treat the absence of a label as indication that the image is a negative example for that label.
Mechanical Turk experiments described in \autoref{sec:mech_turk_evaluation} serve to allay our concerns.

\paragraph{Wikipaintings.}
We also provide a new dataset for classifying painting style.
To our knowledge, no previous large-scale dataset exists for this task -- although very recently a large dataset of artwork did appear for other tasks \cite{Mensink2014}.
We collect a dataset of 100,000 high-art images -- mostly paintings -- labeled with artist, style, genre, date, and free-form tag information by a community of experts on the \texttt{Wikipaintings.org} website.

Analyzing style of non-photorealistic media is an interesting problem, as much of our present understanding of visual style arises out of thousands of years of developments in fine art, marked by distinct historical styles.
Our dataset presents significant stylistic diversity, primarily spanning Renaissance styles to modern art movements (\autoref{fig:wikipaintings_data} provides further breakdowns).
We select 25 styles with more than 1,000 examples, for a total of 85,000 images.
Example images are shown in~\autoref{fig:wikipaintings_style_examples}.
 
    \makeatletter{}\section{Learning algorithm}

We learn to classify novel images according to their style, using the labels assembled in the previous section.
Because the datasets we deal with are quite large and some of the features are high-dimensional, we consider only linear classifiers, relying on sophisticated features to provide robustiness.

We use an open-source implementation of Stochastic Gradient Descent with adaptive subgradient \cite{Agarwal-JMLR-2012}.
The learning process optimizes the function \[
\underset{w}{\text{min }} \lambda_1 \|w\|_1 + \frac{\lambda_2}{2} \Vert w \Vert_2^2 + \sum_i \ell(x_i, y_i, w)
\]
We set the $L_1$ and $L_2$ regularization parameters and the form of the loss function by validation on a held-out set.
For the loss $\ell(x, y, w)$, we consider the hinge ($\max(0, 1 - y \cdot w^T x)$) and logistic ($\log(1 + \exp(-y \cdot w^T x))$) functions.
We set the initial learning rate to $0.5$, and use adaptive subgradient optimization~\cite{duchi2011adaptive}.
Our setup is of multi-class classification; we use the One vs. All reduction to binary classifiers.
 
    \makeatletter{}\section{Image Features}

In order to classify styles, we must choose appropriate image features.  We hypothesize that image style may be related to many different features, including low-level statistics \cite{Lyu-PNAS-2004}, color choices, composition, and content.  Hence, we test features that embody these different elements, including features from the object recognition literature.
We evaluate single-feature performance, as well as second-stage fusion of multiple features.

\vspace{-1em}
\paragraph{L*a*b color histogram.}
Many of the Flickr styles exhibit strong dependence on color. For example, \emph{Noir} images are nearly all black-and-white, while most \emph{Horror} images are very dark, and \emph{Vintage} images use old photographic colors. We use a standard color histogram feature, computed on the whole image.
The 784-dimensional joint histogram in CIELAB color space has 4, 14, and 14 bins in the L*, a*, and b* channels, following Palermo et al.~\cite{Palermo-ECCV-2012}, who showed this to be the best performing single feature for determining the date of historical color images.

\vspace{-1em}
\paragraph{GIST.}
The classic gist descriptor \cite{Oliva-IJCV-2001} is known to perform well for scene classification and retrieval of images visually similar at a low-resolution scale, and thus can represent image composition to some extent.
We use the INRIA LEAR implementation, resizing images to 256 by 256 pixels and extracting a 960-dimensional color GIST feature.

\vspace{-1em}
\paragraph{Graph-based visual saliency.}
We also model composition with a visual attention feature \cite{Harel-NIPS-2006}.
The feature is fast to compute and has been shown to predict human fixations in natural images basically as well as an individual human (humans are far better in aggregate, however).
The 1024-dimensional feature is computed from images resized to 256 by 256 pixels.

\vspace{-1em}
\paragraph{Meta-class binary features.}
Image content can be predictive of individual styles, e.g., \emph{Macro} images include many images of insects and flowers. The \texttt{mc-bit} feature~\cite{Bergamo-CVPR-2012} is a 15,000-dimensional bit vector feature learned as a non-linear combination of classifiers trained using existing features (e.g., SIFT, GIST, Self-Similarity) on thousands of random ImageNet synsets, including internal ILSVRC2010 nodes.
In essence, MC-bit is a hand-crafted ``deep'' architecture, stacking classifiers and pooling operations on top of lower-level features.

\vspace{-1em}
\paragraph{Deep convolutional net.}
Current state-of-the-art results on ImageNet, the largest image classification challenge, have come from a deep convolutional network trained in a fully-supervised manner \cite{krizhevsky2012imagenet}.
We use the Caffe \cite{Jia13caffe} open-source implementation of the ImageNet-winning eght-layer convolutional network, trained on over a million images annotated with 1,000 ImageNet classes.
We investigate using features from two different levels of the network, referred to as DeCAF$_5$ and DeCAF$_6$ (following \cite{Donahue2013}).
The features are 8,000- and 4,000-dimensional and are computed from images center-cropped and resized to 256 by 256 pixels.

\vspace{-1em}
\paragraph{Content classifiers.}
Following Dhar et al.~\cite{Dhar-CVPR-2011}, who use high-level classifiers as features for their aesthetic rating prediction task, we evaluate using object classifier confidences as features.
Specifically, we train classifiers for all 20 classes of the PASCAL VOC \cite{pascal-voc-2010} using the DeCAF$_6$ feature.
The resulting classifiers are quite reliable, obtaining $0.7$ mean AP on the VOC 2012.

We aggregate the data to train four classifiers for ``animals'', ``vehicles'', ``indoor objects'' and ``people''.
These aggregate classes are presumed to discriminate between vastly different types of images -- types for which different style signals may apply.
For example, a \emph{Romantic} scene with people may be largely about the composition of the scene, whereas, \emph{Romantic} scenes with vehicles may be largely described by color.

To enable our classifiers to learn content-dependent style, we can take the outer product of a feature channel with the four aggregate content classifiers.
 
    \makeatletter{}\section{Experiments}

\makeatletter{}\begin{table}
\caption{
    Mean APs on three datasets for the considered single-channel features and their second-stage combination.
    As some features were clearly worse than others on the AVA Style dataset, only the better features were evaluated on larger datasets.
}
\label{tab:mean_aps}
\centering
\vspace{1em}
\footnotesize{
\begin{tabular}{lrrrrrrrrr}
\toprule
{}                & Fusion x Content & DeCAF$_6$ & MC-bit & L*a*b* Hist & GIST  & Saliency & random \\
\midrule
AVA Style         & \textbf{0.581}   & 0.579     & 0.539  & 0.288       & 0.220 & 0.152    & 0.132 \\
Flickr            & \textbf{0.368}   & 0.336     & 0.328  & -           & -     & -        & 0.052 \\
Wikipaintings     & \textbf{0.473}   & 0.356     & 0.441  & -           & -     & -        & 0.043 \\
\bottomrule
\end{tabular}
}
\end{table}

\subsection{Flickr Style}
We learn and predict style labels on the 80,000 images labeled with 20 different visual styles of our new Flickr Style dataset, using 20\% of the data for testing, and another 20\% for parameter-tuning validation.

There are several performance metrics we consider.
Average Precision evaluation (as reported in \autoref{tab:mean_aps} and in \autoref{tab:flickr_ap_table}) is computed on a random class-balanced subset of the test data (each class has equal prevalence).
We compute confusion matrices (\autoref{fig:flickr_conf}, \autoref{fig:wp_conf}, \autoref{fig:ava_style_conf}) on the same data.
Per-class accuracies are computed on subsets of the data balanced by the binary label, such that chance performance is 50\%.
We follow these decisions in all following experiments.

The best single-channel feature is DeCAF$_6$ with 0.336 mean AP; feature fusion obtains 0.368 mean AP.
Per-class APs range from 0.17 [Depth of Field] to  0.62 [Macro].
Per-class accuracies range from 68\% [Romantic, Depth of Field] to 85\% [Sunny, Noir, Macro].
The average per-class accuracy is 78\%.
We show the most confident style classifications on the test set of Flickr Style in \autoref{fig:flickr_on_flickr}.

Upon inspection of the confusion matrices, we saw points of understandable confusion: Depth of Field vs.~Macro, Romantic vs.~Pastel, Vintage vs.~Melancholy.
There are also surprising sources of mistakes: Macro vs.~Bright/Energetic, for example.
To explain this particular confusion, we observed that lots of Macro photos contain bright flowers, insects, or birds, often against vibrant greenery.
Here, at least, the content of the image dominates its style label.

To explore further content-style correlations, we plot the outputs of PASCAL object class classifiers (one of our features) on the Flickr dataset in~\autoref{fig:content_correlation}.
We can observe that some styles have strong correlations to content (e.g., ``Hazy'' occurs with ``vehicle'', ``HDR'' doesn't occur with ``cat'').

We hypothesize that style is content-dependent: a Romantic portrait may have different low-level properties than a Romantic sunset. We form a new feature as an outer product of our content classifier features with the second-stage late fusion features (``Fusion $\times$ Content'' in all results figures).  These features gave the best results, thus supporting the hypothesis.

\begin{figure}[h]
\centering
    \includegraphics[width=\linewidth]{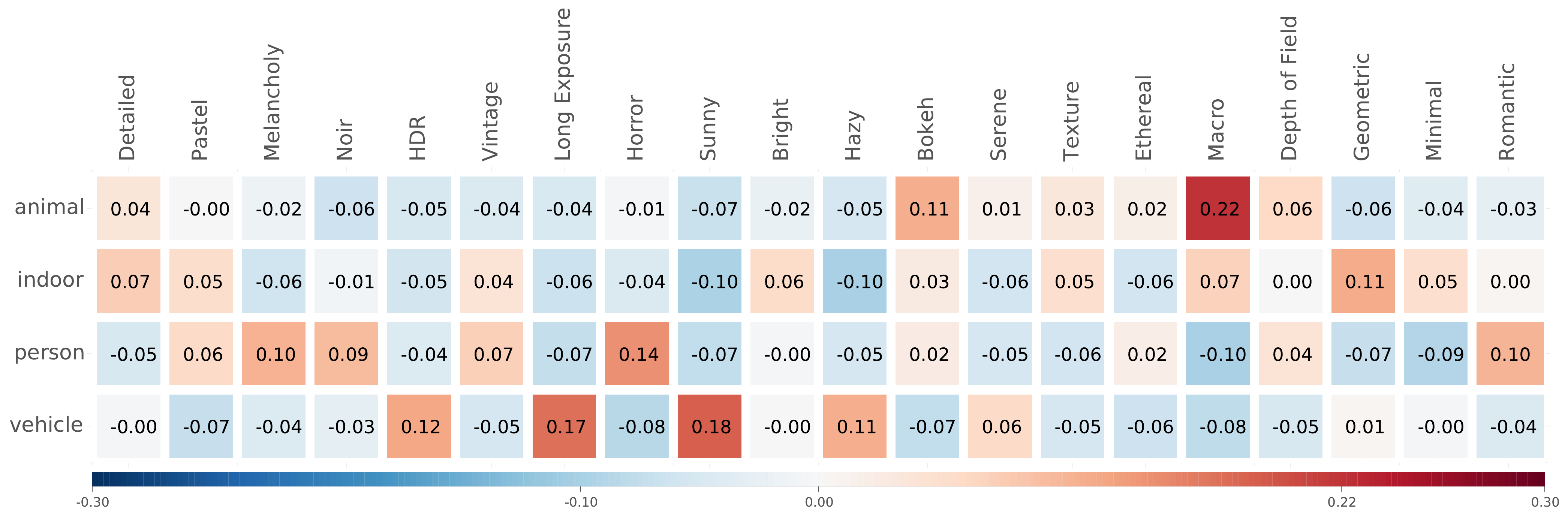}
    \caption{
        Correlation of PASCAL content classifier predictions (rows) against ground truth Flickr Style labels (columns).
        We see, for instance, that the Macro style is highly correlated with presence of animals, and that Long Exposure and Sunny style photographs often feature vehicles.
    }\label{fig:content_correlation}
\end{figure}

\vspace{-.5em}
\paragraph{Mechanical Turk Evaluation.}\label{sec:mech_turk_evaluation}
In order to provide a human baseline for evaluation, we performed a Mechanical Turk study.
For each style, Turkers were shown positive and negative examples for each Flickr Group, and then they evaluated whether each image in the test set was part of the given style.
We treat the Flickr group memberships as ground truth as before, and then evaluate Turkers' ability to accurately determine group membership.
Measures were taken to remove spam workers; see \autoref{sec:mech_turk_details} for our experimental setup.
For efficiency, one quarter of the test set was used, and two redundant styles (Bokeh and Detailed) were removed.
Each test image was evaluated by 3 Turkers, and the majority vote taken as the human result for this image.

Results are presented in \autoref{tab:flickr_vs_mturk}.
In total, Turkers achieved 75\% mean accuracy (ranging from 61\% [Romantic] to 92\% [Macro]) across styles, in comparison to 78\% mean accuracy (ranging from 68\% [Depth of Field] to 87\% [Macro]) of our best method.
Our algorithm did significantly worse than Turkers on Macro and Horror, and significantly better on Vintage, Romantic, Pastel, Detailed, HDR, and Long Exposure styles.

Some of this variance may be due to subtle difference from the Turk tasks that we provided, as compared to the definitions of the Flickr groups, but may also due to the Flickr groups' incorporating images that do not quite fit the common definition of the given style.
For example, there may be a mismatch between different notions of ``romantic'' and ``vintage,'' and how inclusively these terms are defined.

We additionally used the Turker opinion as ground truth for our method's predictions.
In switching from the default Flickr to the MTurk ground truth, our method's accuracy hardly changed from 78\% to 77\%.
However, we saw that the accuracy of our Vintage, Detailed, Long Exposure, Minimal, HDR, and Sunny style classifiers significantly decreased, indicating machine-human disagreement on those styles.

\subsection{Wikipaintings}
With the same setup and features as in the Flickr experiments, we evaluate 85,000 images labeled with 25 different art styles.
Detailed results are provided in \autoref{tab:wikipaintings_ap_table} and \autoref{tab:wp_accuracies}.
The best single-channel feature is MC-bit with 0.441 mean AP; feature fusion obtains 0.473 mean AP.
Per-class accuracies range from 72\% [Symbolism, Expressionism, Art Nouveau] to 94\% [Ukiyo-e, Minimalism, Color Field Painting].

\subsection{AVA Style}
AVA \cite{Murray-CVPR-2012} is a dataset of 250K images from \texttt{dpchallenge.net}.
We evaluate classification of aesthetic rating and of 14 different photographic style labels on the 14,000 images of the AVA dataset that have such labels.
For the style labels, the publishers of the dataset provide a train/test split, where training images have only one label, but test images may have more than one label \cite{Murray-CVPR-2012}.
Our results are presented in \autoref{tab:ava_style_ap_table}.
For style classification, the best single feature is the DeCAF$_6$ convolution network feature, obtaining 0.579 mean AP.
Feature fusion improves the result to 0.581 mean AP; both results beat the previous state-of-the-art of 0.538 mean AP \cite{Murray-CVPR-2012}.
\footnote{Our results beat 0.54 mAP using both the AVA-provided class-imbalanced test split, and the class-balanced subsample that we consider to be more correct evaluation, and for which we provide numbers.}

In all metrics, the DeCAF and MC-bit features significantly outperformed more low-level features on this dataset.
Accordingly, we do not evaluate the low-level features on the larger Flickr and Wikipaintings datasets.
 
    \makeatletter{}
Test images were grouped into 10 images per Human Interface Task (HIT). Each task asks the Turker to evaluate the style (e.g., ``Is this image VINTAGE?'') for each image.  For each style, we provided a short blurb describing the style in words, and provided 12-15 hand-chosen positive and negative examples for each Flickr Group.
Each HIT included 2 sentinels: images which were very clearly positives and similar to the examples.  HITs were rejected when Turkers got both sentinels wrong.
Turkers were paid $0.10$ per HIT, and were allowed to perform multiple hits.  Manual inspection of the results indicate that the Turkers understood the task and were performing effectively.  A few Turkers sent unsolicited feedback indicating that they were really enjoying the HITs (``some of the photos are beautiful'') and wanted to perform them as effectively as possible.
 
    \makeatletter{}\subsection{Application: Style-Based Image Search}

\makeatletter{}\newcommand{\fgap}{.6in}
\begin{figure}[h!]
\centering
\begin{tabular}{m{.02in}|m{\fgap} m{\fgap} m{\fgap} m{\fgap} m{\fgap}}
    \begin{turn}{90}\footnotesize{Minimal}\end{turn} &
    \includegraphics[width=.75in]{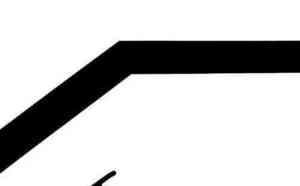} &
    \includegraphics[width=.75in]{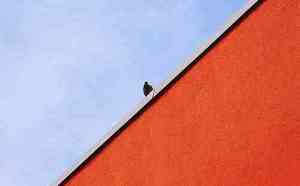} &
    \includegraphics[width=.75in]{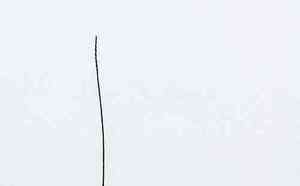} &
    \includegraphics[width=.75in]{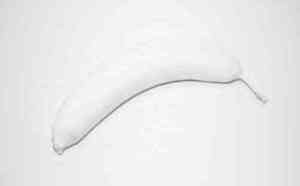} &
    \includegraphics[width=.75in]{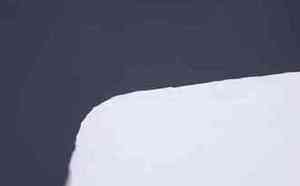} \\
    \begin{turn}{90}\footnotesize{Melancholy}\end{turn} &
    \includegraphics[width=.75in]{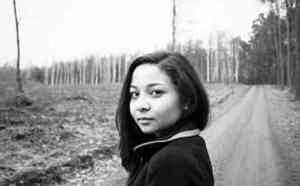} &
    \includegraphics[width=.75in]{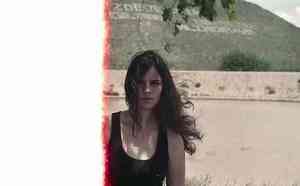} &
    \includegraphics[width=.75in]{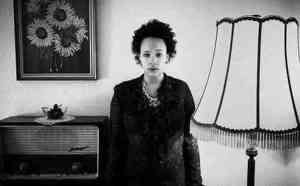} &
    \includegraphics[width=.75in]{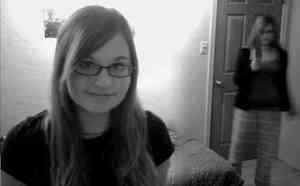} &
    \includegraphics[width=.75in]{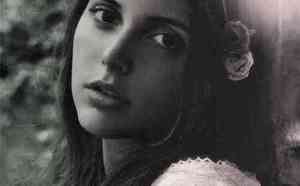} \\
    \begin{turn}{90}\footnotesize{HDR}\end{turn} &
    \includegraphics[width=.75in]{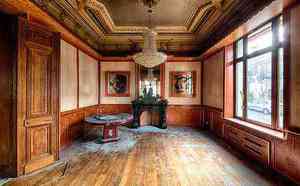} &
    \includegraphics[width=.75in]{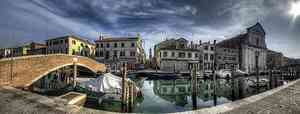} &
    \includegraphics[width=.75in]{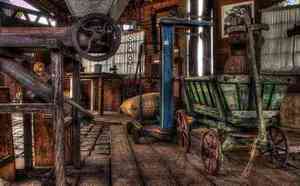} &
    \includegraphics[width=.75in]{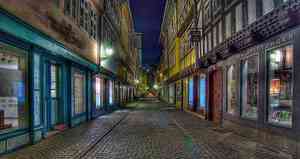} &
    \includegraphics[width=.75in]{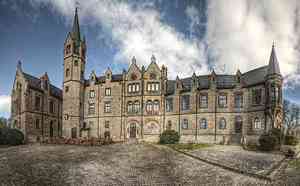} \\
    \begin{turn}{90}\footnotesize{Vintage}\end{turn} &
    \includegraphics[width=.75in]{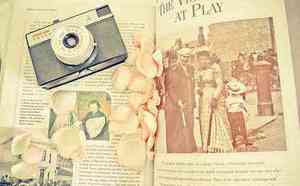} &
    \includegraphics[width=.75in]{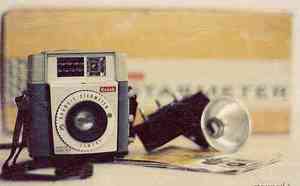} &
    \includegraphics[width=.75in]{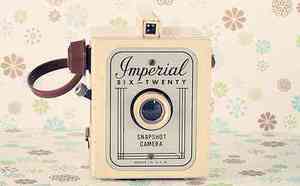} &
    \includegraphics[width=.75in]{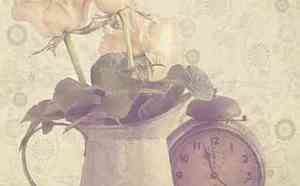} &
    \includegraphics[width=.75in]{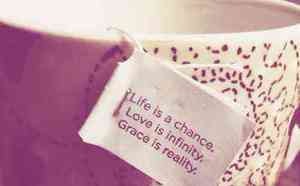} \\
\end{tabular}
\vspace{1em}
\caption{
    Top five most-confident positive predictions on the Flickr Style test set, for a few different styles.
    }\label{fig:flickr_on_flickr}
\end{figure}

Style classifiers learned on our datasets can be used toward novel goals.
For example, sources of stock photography or design inspiration may be better navigated with a vocabulary of style.
Currently, companies expend labor to manually annotate stock photography with such labels.
With our approach, any image collection can be searchable and rankable by style.

To demonstrate, we apply our Flickr-learned style classifiers to a new dataset of 80K images gathered on Pinterest (also available with our code release); some results are shown in \autoref{fig:flickr_on_pinterest}.
Interestingly, styles learned from photographs can be used to order paintings, and styles learned from paintings can be used to order photographs, as illustrated in \autoref{fig:photo_painting}.

\begin{figure*}[ht]
\centering
\includegraphics[width=.95\linewidth]{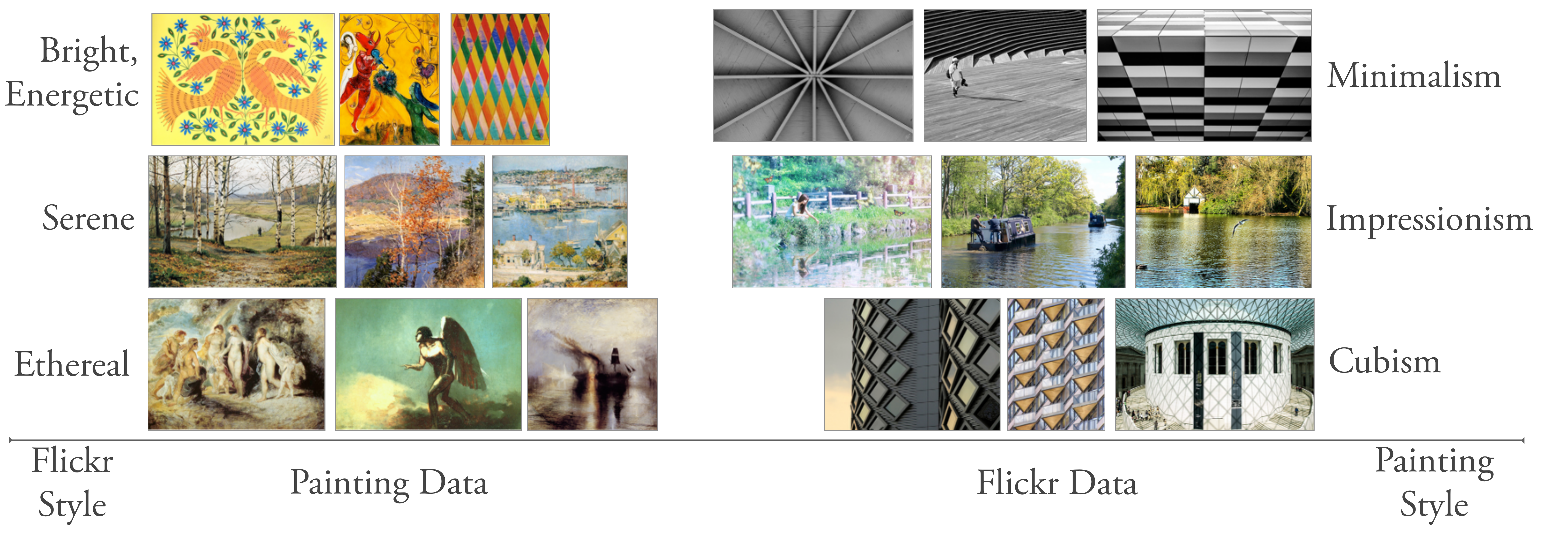}
\vspace{-1ex}
\caption{
    Cross-dataset style.
    On the left are shown top scorers from the Wikipaintings set, for styles learned on the Flickr set.
    On the right, Flickr photographs are accordingly sorted by Painting style.
    (Figure best viewed in color.)
}
\label{fig:photo_painting}
\end{figure*}

\makeatletter{}\newcommand{\dgap}{.42in}
\begin{figure}
\begin{subfigure}[t]{0.48\linewidth}
    \begin{tabular}{m{.02in}|m{\dgap} m{\dgap} m{\dgap}}
    \begin{turn}{90}\small{Bright}\end{turn} &
    \includegraphics[width=.53in]{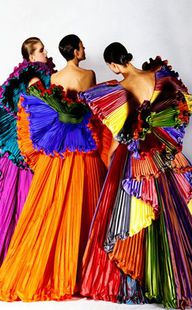} &
    \includegraphics[width=.53in]{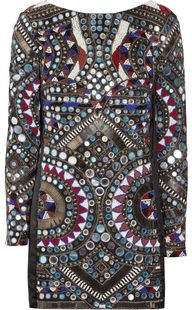} &
    \includegraphics[width=.53in]{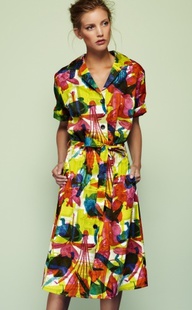} \\ \\
    \begin{turn}{90}\small{Pastel}\end{turn} &
    \includegraphics[width=.53in]{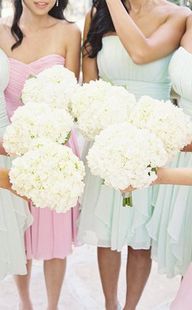} &
    \includegraphics[width=.53in]{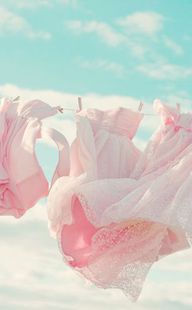} &
    \includegraphics[width=.53in]{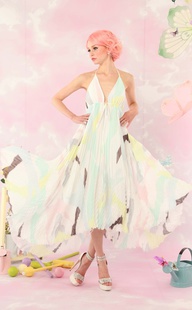} \\ \\
    \begin{turn}{90}\small{Ethereal}\end{turn} &
    \includegraphics[width=.53in]{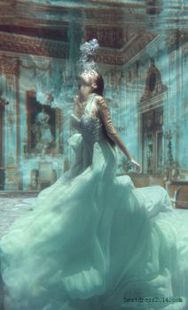} &
    \includegraphics[width=.53in]{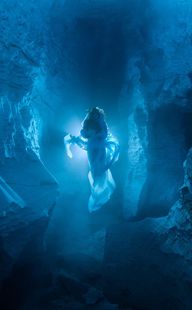} &
    \includegraphics[width=.53in]{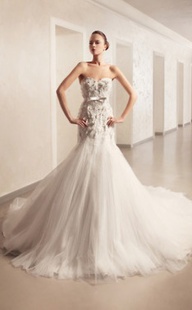} \\ \\
    \begin{turn}{90}\small{Noir}\end{turn} &
    \includegraphics[width=.53in]{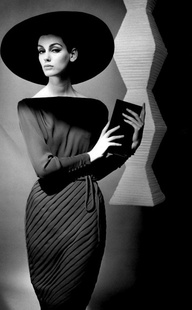} &
    \includegraphics[width=.53in]{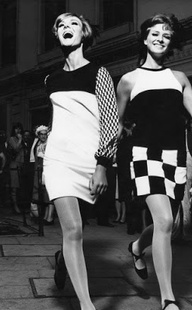} &
    \includegraphics[width=.53in]{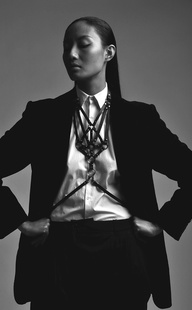} \\ \\
    \begin{turn}{90}\small{Vintage}\end{turn} &
    \includegraphics[width=.53in]{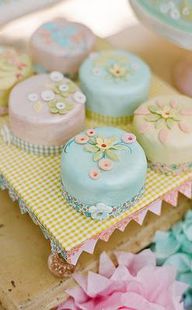} &
    \includegraphics[width=.53in]{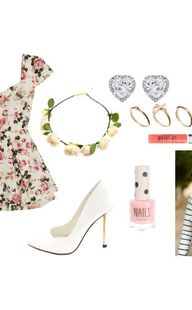} &
    \includegraphics[width=.53in]{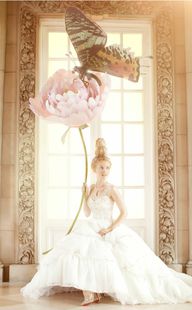} \\
    \end{tabular}
    \caption{Query: ``dress''.}
\end{subfigure}\hfill\begin{subfigure}[t]{0.48\linewidth}
    \begin{tabular}{m{.02in}|m{\dgap} m{\dgap} m{\dgap}}
    \begin{turn}{90}\small{DoF}\end{turn} &
    \includegraphics[width=.53in]{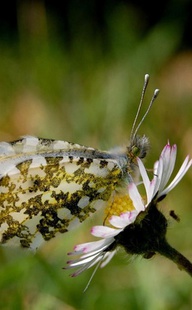} &
    \includegraphics[width=.53in]{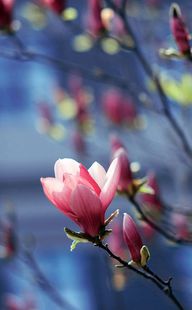} &
    \includegraphics[width=.53in]{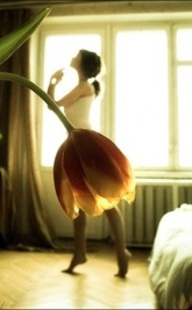} \\ \\
    \begin{turn}{90}\small{Romantic}\end{turn} &
    \includegraphics[width=.53in]{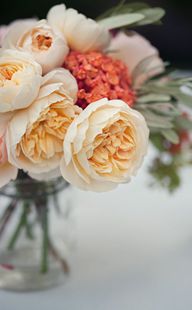} &
    \includegraphics[width=.53in]{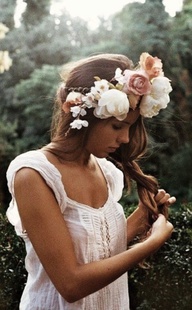} &
    \includegraphics[width=.53in]{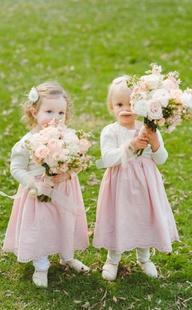} \\ \\
    \begin{turn}{90}\small{Sunny}\end{turn} &
    \includegraphics[width=.53in]{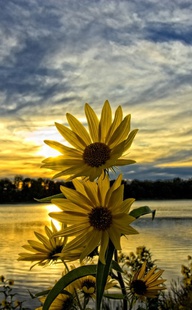} &
    \includegraphics[width=.53in]{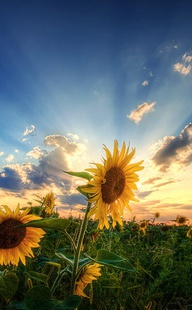} &
    \includegraphics[width=.53in]{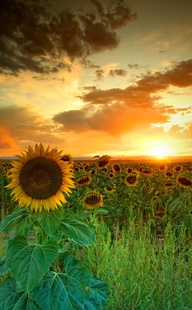} \\ \\
    \begin{turn}{90}\small{Geometric}\end{turn} &
    \includegraphics[width=.53in]{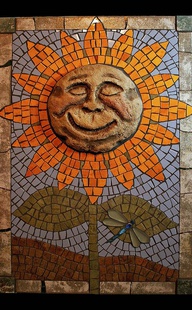} &
    \includegraphics[width=.53in]{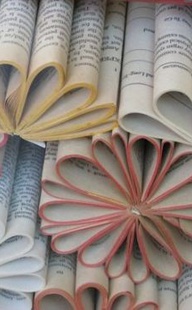} &
    \includegraphics[width=.53in]{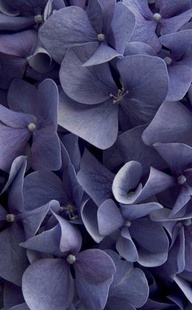} \\ \\
    \begin{turn}{90}\small{Serene}\end{turn} &
    \includegraphics[width=.53in]{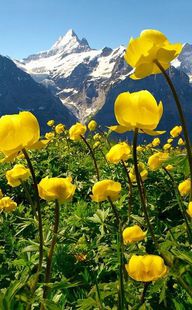} &
    \includegraphics[width=.53in]{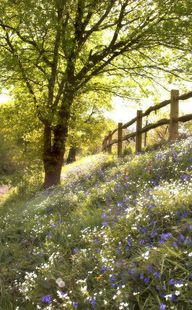} &
    \includegraphics[width=.53in]{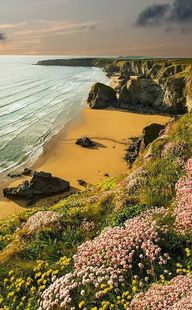} \\
    \end{tabular}
    \caption{Query: ``flower''.}
\end{subfigure}
\\
\caption{
    Example of filtering image search results by style.
    Our Flickr Style classifiers are applied to images found on Pinterest.
    The images are searched by the text contents of their captions, then filtered by the response of the style classifiers.
    Here we show three out of top five results for different query/style combinations.
}\label{fig:flickr_on_pinterest}
\end{figure}

    \makeatletter{}\subsection{Discussion}

We have made significant progress in defining the problem of understanding photographic style.
We provide a novel dataset that exhibits several types of styles not previously considered in the literature, and we demonstrate state-of-the-art results in prediction of both style and aesthetic quality.
These results are comparable to human performance.
We also show that style is highly content-dependent.

Style plays a significant role in much of the manmade imagery we experience daily, and there is considering need for future work to further answer the question ``What is style?''

One of the most interesting outcomes of this work is the success of features trained for object detection for both aesthetic and style classification.
We propose several possible hypotheses to explain these results.
Perhaps the network layers that we use as features are extremely good as general visual features for image representation in general.
Another explanation is that object recognition depends on object appearance, e.g., distinguishing red from white wine, or different kinds of terriers, and that the model learns to repurpose these features for image style.
Understanding and improving on these results is fertile ground for future work.

    {\footnotesize
    \bibsep=3pt

    }
\fi

            \makeatletter{}\begin{table}
\centering
\caption{Exact Flickr group names, and their sizes.}\label{tab:flickr_groups}
\vspace{1em}
\begin{tabular}{rl}
    \textbf{Style}        & \textbf{Group names [num images]} \\
    \midrule
    Bokeh                 & Bokeh Photography (1/day) [187K] \\
    Bright                & Colour Mania [100K] \\
    Depth of Field        & Depth of Field [116K], Finest DoF [54K] \\
    Detailed              & Details aller Art - Details of all kind [22K], Detail pictures [5K] \\
    Ethereal              & Ethereal World [21K] \\
    Geometric Composition & Geometric Beauty [168K] \\
    Hazy                  & Misty hazy smokey [14K] \\
    HDR                   & HDR ADDICTED [374K]\\
    Horror                & Horror [16K] \\
    Long Exposure         & Long Exposure [619K] \\
    Macro                 & Closer and Closer Macro Photography [990K] \\
    Melancholy            & melancholy [106K] \\
    Minimal               & Less Is More... [44K] \\
    Noir                  & Film Noir Mood [7K] \\
    Romantic              & Romantic Images [20K] \\
    Serene                & \~ Serene \~ [68K] \\
    Pastel                & pastel and dreamy [120K], Pastel Soft tone [7K] \\
    Sunny                 & Sun, sun and more sun [23K] \\
    Texture               & Texture [103K] \\
    Vintage               & Vintage Feelings [4K], Vintage \& Retro [61K] \\
    \bottomrule
\end{tabular}
\end{table}
 
    \makeatletter{}\begin{figure}[th]
\centering
\includegraphics[width=\linewidth]{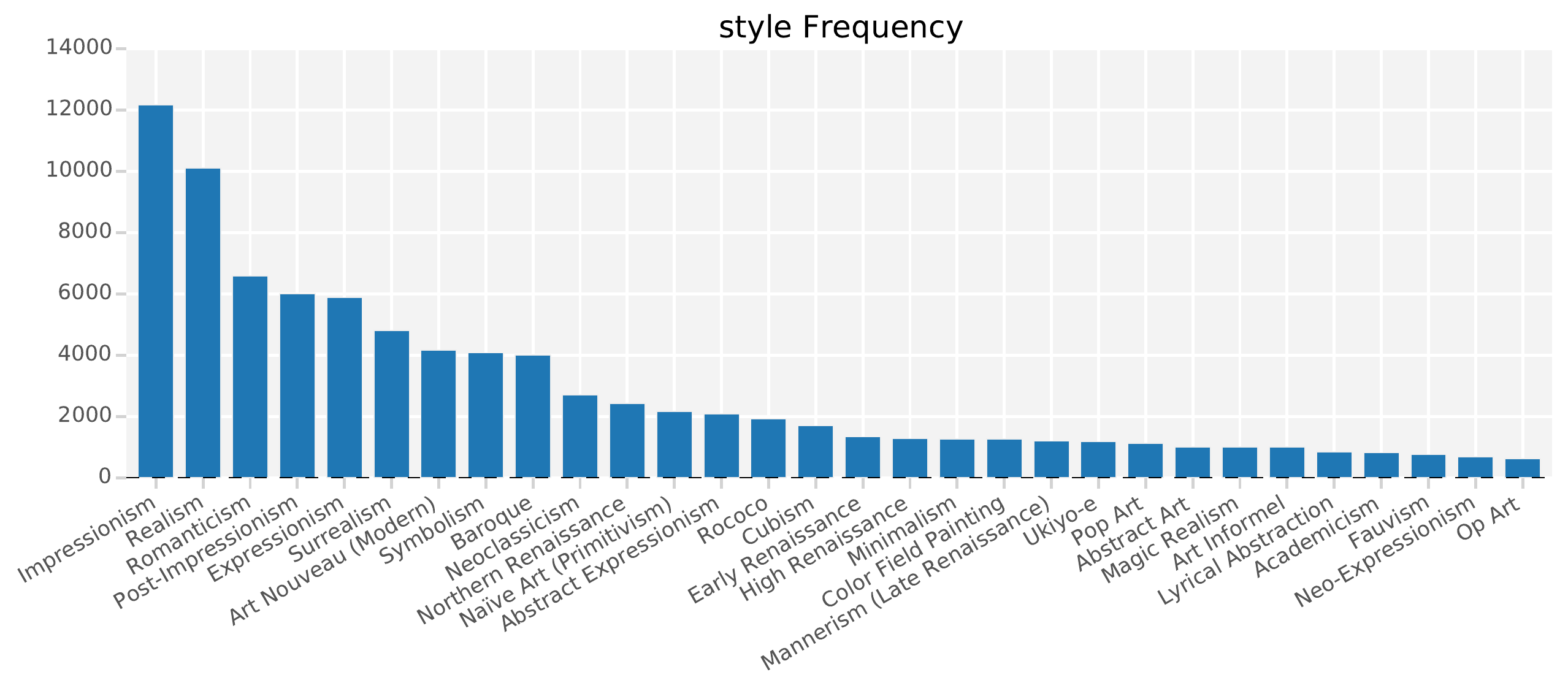}\\
\includegraphics[width=\linewidth]{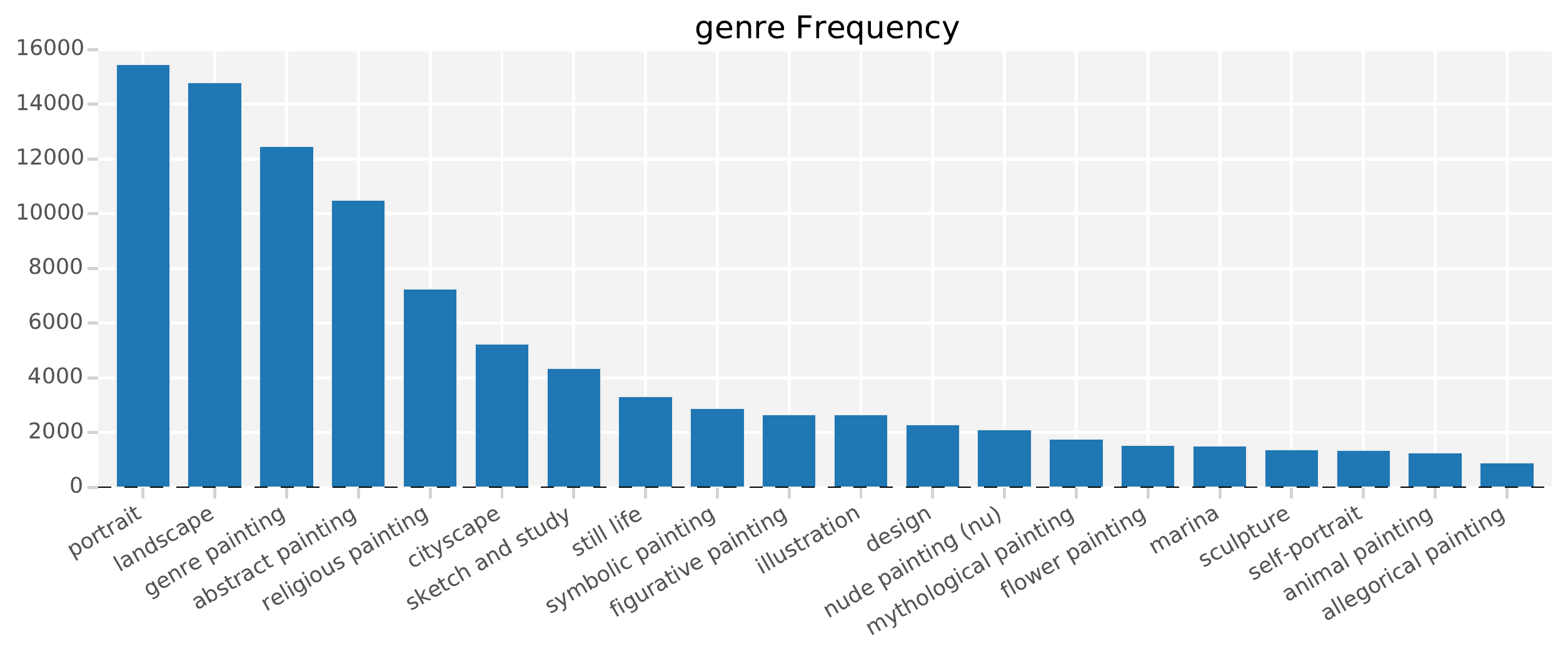}\\
\includegraphics[width=\linewidth]{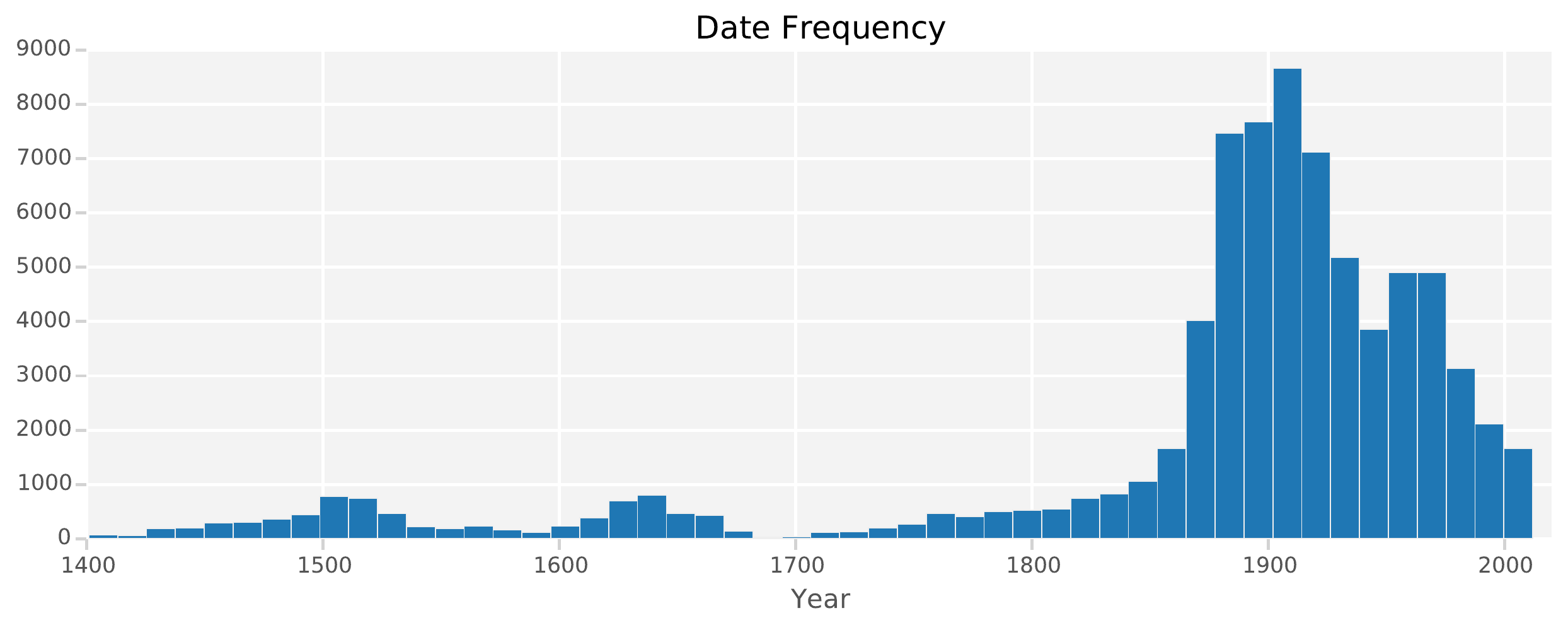}
\caption{Distribution of image style, genre, and date in the Wikipaintings dataset.}
\label{fig:wikipaintings_data}
\end{figure}
 
    \makeatletter{}\begin{table*}[h!]
\caption{
    All per-class APs on all evaluated features on the AVA Style dataset.
}\label{tab:ava_style_ap_table}
\vspace{1em}
\centering
\small{
\makeatletter{}\begin{tabular}{lllllrlllll}
\toprule
{}                    & Fusion & DeCAF$_6$ & MC-bit & Murray & L*a*b* & GIST  & Saliency \\
\midrule
Complementary\_Colors & 0.469  & 0.548     & 0.329  & 0.440  & 0.294  & 0.223 & 0.111 \\
Duotones              & 0.676  & 0.737     & 0.612  & 0.510  & 0.582  & 0.255 & 0.233 \\
HDR                   & 0.669  & 0.594     & 0.624  & 0.640  & 0.194  & 0.124 & 0.101 \\
Image\_Grain          & 0.647  & 0.545     & 0.744  & 0.740  & 0.213  & 0.104 & 0.104 \\
Light\_On\_White      & 0.908  & 0.915     & 0.802  & 0.730  & 0.867  & 0.704 & 0.172 \\
Long\_Exposure        & 0.453  & 0.431     & 0.420  & 0.430  & 0.232  & 0.159 & 0.147 \\
Macro                 & 0.478  & 0.427     & 0.413  & 0.500  & 0.230  & 0.269 & 0.161 \\
Motion\_Blur          & 0.478  & 0.467     & 0.458  & 0.400  & 0.117  & 0.114 & 0.122 \\
Negative\_Image       & 0.595  & 0.619     & 0.499  & 0.690  & 0.268  & 0.189 & 0.123 \\
Rule\_of\_Thirds      & 0.352  & 0.353     & 0.236  & 0.300  & 0.188  & 0.167 & 0.228 \\
Shallow\_DOF          & 0.624  & 0.659     & 0.637  & 0.480  & 0.332  & 0.276 & 0.223 \\
Silhouettes           & 0.791  & 0.801     & 0.801  & 0.720  & 0.261  & 0.263 & 0.130 \\
Soft\_Focus           & 0.312  & 0.354     & 0.290  & 0.390  & 0.127  & 0.126 & 0.114 \\
Vanishing\_Point      & 0.684  & 0.658     & 0.685  & 0.570  & 0.123  & 0.107 & 0.161 \\
\midrule
mean                  & 0.581  & 0.579     & 0.539  & 0.539  & 0.288  & 0.220 & 0.152 \\
\bottomrule
\end{tabular}
 
}
\end{table*}

\begin{table*}[h!]
\caption{
    All per-class APs on all evaluated features on the Flickr dataset.
}\label{tab:flickr_ap_table}
\vspace{1em}
\centering
\makeatletter{}\begin{tabular}{llll}
\toprule
{}                     & Fusion x Content & DeCAF$_6$ & MC-bit \\
\midrule
Bokeh                  & 0.288            & 0.253     & 0.248 \\
Bright                 & 0.251            & 0.236     & 0.183 \\
Depth\_of\_Field       & 0.169            & 0.152     & 0.148 \\
Detailed               & 0.337            & 0.277     & 0.278 \\
Ethereal               & 0.408            & 0.393     & 0.335 \\
Geometric\_Composition & 0.411            & 0.355     & 0.360 \\
HDR                    & 0.487            & 0.406     & 0.475 \\
Hazy                   & 0.493            & 0.451     & 0.447 \\
Horror                 & 0.400            & 0.396     & 0.295 \\
Long\_Exposure         & 0.515            & 0.457     & 0.463 \\
Macro                  & 0.617            & 0.582     & 0.530 \\
Melancholy             & 0.168            & 0.147     & 0.136 \\
Minimal                & 0.512            & 0.444     & 0.481 \\
Noir                   & 0.494            & 0.481     & 0.408 \\
Pastel                 & 0.258            & 0.245     & 0.211 \\
Romantic               & 0.227            & 0.204     & 0.185 \\
Serene                 & 0.281            & 0.257     & 0.239 \\
Sunny                  & 0.500            & 0.481     & 0.453 \\
Texture                & 0.265            & 0.227     & 0.229 \\
Vintage                & 0.282            & 0.273     & 0.222 \\
\midrule
mean                   & 0.368            & 0.336     & 0.316 \\
\bottomrule
\end{tabular}
 
\end{table*}

\begin{table*}[h!]
\centering
\caption{
    All per-class APs on all evaluated features on the Wikipaintings dataset.
}\label{tab:wikipaintings_ap_table}
\vspace{1em}
\makeatletter{}
\begin{tabular}{llllll}
\toprule
{}                             & Fusion x Content & MC-bit & DeCAF$_6$  \\
\midrule
Abstract\_Art                  & 0.341            & 0.314  & 0.258      \\
Abstract\_Expressionism        & 0.351            & 0.340  & 0.243      \\
Art\_Informel                  & 0.221            & 0.217  & 0.187      \\
Art\_Nouveau\_(Modern)         & 0.421            & 0.402  & 0.197      \\
Baroque                        & 0.436            & 0.386  & 0.313      \\
Color\_Field\_Painting         & 0.773            & 0.739  & 0.689      \\
Cubism                         & 0.495            & 0.488  & 0.400      \\
Early\_Renaissance             & 0.578            & 0.559  & 0.453      \\
Expressionism                  & 0.235            & 0.230  & 0.186      \\
High\_Renaissance              & 0.401            & 0.345  & 0.288      \\
Impressionism                  & 0.586            & 0.528  & 0.411      \\
Magic\_Realism                 & 0.521            & 0.465  & 0.428      \\
Mannerism\_(Late\_Renaissance) & 0.505            & 0.439  & 0.356      \\
Minimalism                     & 0.660            & 0.614  & 0.604      \\
Nave\_Art\_(Primitivism)       & 0.395            & 0.425  & 0.225      \\
Neoclassicism                  & 0.601            & 0.537  & 0.399      \\
Northern\_Renaissance          & 0.560            & 0.478  & 0.433      \\
Pop\_Art                       & 0.441            & 0.398  & 0.281      \\
Post-Impressionism             & 0.348            & 0.348  & 0.292      \\
Realism                        & 0.408            & 0.309  & 0.266      \\
Rococo                         & 0.616            & 0.548  & 0.467      \\
Romanticism                    & 0.392            & 0.389  & 0.343      \\
Surrealism                     & 0.262            & 0.247  & 0.134      \\
Symbolism                      & 0.390            & 0.390  & 0.260      \\
Ukiyo-e                        & 0.895            & 0.894  & 0.788      \\
\midrule
mean                           & 0.473            & 0.441  & 0.356      \\
\bottomrule
\end{tabular}
 
\end{table*}

\begin{table*}[h!]
\caption
[Comparison of Flickr Style per-class accuracies for our method and Mech Turkers.]
{
Comparison of Flickr Style per-class accuracies for our method and Mech Turkers.
We first give the full results table, then show the signficant deviations between human and machine performance, and between using Flickr and MTurk ground truth.
}\label{tab:flickr_vs_mturk}
\vspace{1em}
\centering
\small{
\makeatletter{}\begin{tabular}{rccc}
\toprule
{}                    & MTurk acc., Flickr g.t. & Our acc., Flickr g.t. & Our acc., MTurk g.t. \\
\midrule
Bright                & 69.10                       & 73.38                     & 73.63 \\
Depth of Field        & 68.92                       & 68.50                     & 81.05 \\
Detailed              & 65.47                       & 75.25                     & 68.44 \\
Ethereal              & 76.92                       & 80.62                     & 77.95 \\
Geometric Composition & 81.52                       & 77.75                     & 80.31 \\
HDR                   & 71.84                       & 82.00                     & 76.96 \\
Hazy                  & 83.49                       & 80.75                     & 81.64 \\
Horror                & 89.85                       & 84.25                     & 81.64 \\
Long Exposure         & 73.12                       & 84.19                     & 76.79 \\
Macro                 & 92.25                       & 86.56                     & 88.39 \\
Melancholy            & 67.77                       & 70.88                     & 71.25 \\
Minimal               & 79.71                       & 83.75                     & 78.57 \\
Noir                  & 81.35                       & 85.25                     & 85.88 \\
Pastel                & 66.94                       & 74.56                     & 75.47 \\
Romantic              & 60.91                       & 68.00                     & 66.25 \\
Serene                & 69.49                       & 70.44                     & 76.80 \\
Sunny                 & 84.48                       & 84.56                     & 79.94 \\
Vintage               & 68.77                       & 75.50                     & 67.80 \\
\midrule
Mean                  & 75.11                       & 78.12                     & 77.15 \\
\end{tabular}

\vfill

\begin{tabular}{rccc}
\toprule
{}                    & Our acc., Flickr g.t.   & Our acc., MTurk g.t.  & \% change from Flickr to MTurk g.t. \\
\midrule
Vintage               & 75.50                       & 67.80                     & -10.19 \\
Detailed              & 75.25                       & 68.44                     & -9.05 \\
Long Exposure         & 84.19                       & 76.79                     & -8.79 \\
Minimal               & 83.75                       & 78.57                     & -6.18 \\
HDR                   & 82.00                       & 76.96                     & -6.15 \\
Sunny                 & 84.56                       & 79.94                     & -5.46 \\
Serene                & 70.44                       & 76.80                     & 9.03 \\
Depth of Field        & 68.50                       & 81.05                     & 18.32 \\
\end{tabular}

\vfill

\begin{tabular}{rccc}
\toprule
{}                     & Our acc., Flickr g.t. & MTurk acc., Flickr g.t. & Acc. difference \\
\midrule
Horror                 & 84.25                     & 90.42                       & -6.17 \\
Macro                  & 86.56                     & 91.71                       & -5.15 \\
Romantic               & 68.00                     & 61.04                       & 6.96 \\
Pastel                 & 74.56                     & 66.87                       & 7.69 \\
HDR                    & 82.00                     & 72.79                       & 9.21 \\
Long Exposure          & 84.19                     & 73.83                       & 10.35 \\
Detailed               & 75.25                     & 63.30                       & 11.95 \\
\end{tabular}
 
}
\end{table*}

\begin{table*}
\caption{
    Per-class accuracies on the Wikipaintings dataset, using the MC-bit feature.
}\label{tab:wp_accuracies}
\vspace{1em}
\centering
\begin{tabular}{lrlr}
\toprule
\textbf{Style} & \textbf{Accuracy} & \textbf{Style} & \textbf{Accuracy} \\
\midrule
Symbolism                    &         71.24 & Impressionism                &         82.15 \\
Expressionism                &         72.03 & Northern Renaissance         &         82.32 \\
Art Nouveau (Modern)         &         72.77 & High Renaissance             &         82.90 \\
Nave Art (Primitivism)       &         72.95 & Mannerism (Late Renaissance) &         83.04 \\
Surrealism                   &         74.44 & Pop Art                      &         83.33 \\
Post-Impressionism           &         74.51 & Early Renaissance            &         84.69 \\
Romanticism                  &         75.86 & Abstract Art                 &         85.10 \\
Realism                      &         75.88 & Cubism                       &         86.85 \\
Magic Realism                &         78.54 & Rococo                       &         87.33 \\
Neoclassicism                &         80.18 & Ukiyo-e                      &         93.18 \\
Abstract Expressionism       &         81.25 & Minimalism                   &         94.21 \\
Baroque                      &         81.45 & Color Field Painting         &         95.58 \\
Art Informel                 &         82.09 &                              &               \\
\bottomrule
\end{tabular}
\end{table*}

\begin{figure*}[h!]
\centering
\includegraphics[width=1\linewidth]{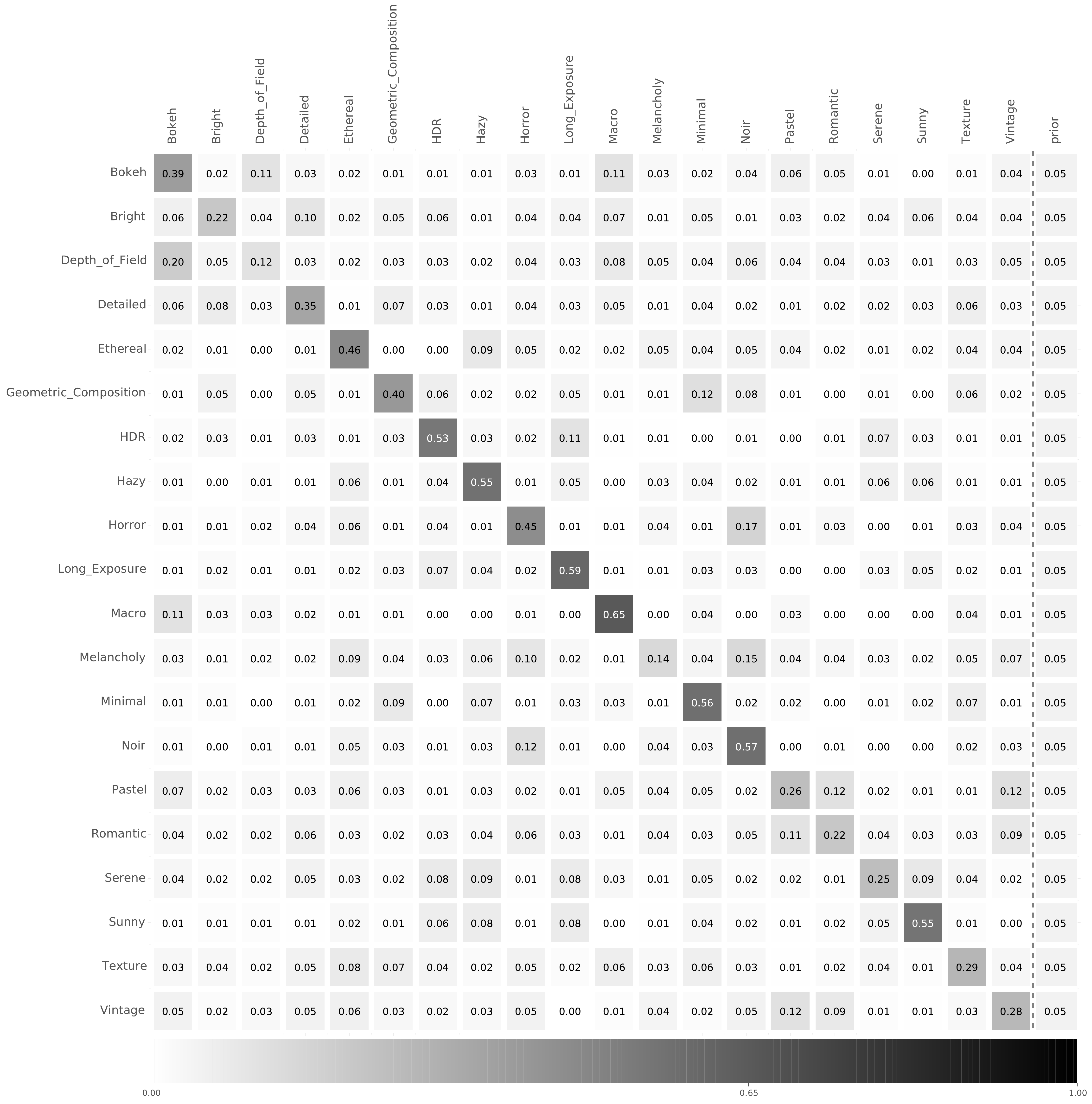}
\caption
[Confusion matrix of our best classifier on the Flickr dataset.]
{Confusion matrix of our best classifier (\mbox{Late-fusion $\times$ Content}) on the Flickr dataset.}
\label{fig:flickr_conf}
\end{figure*}

\begin{figure*}[h!]
\centering
\includegraphics[width=1\linewidth]{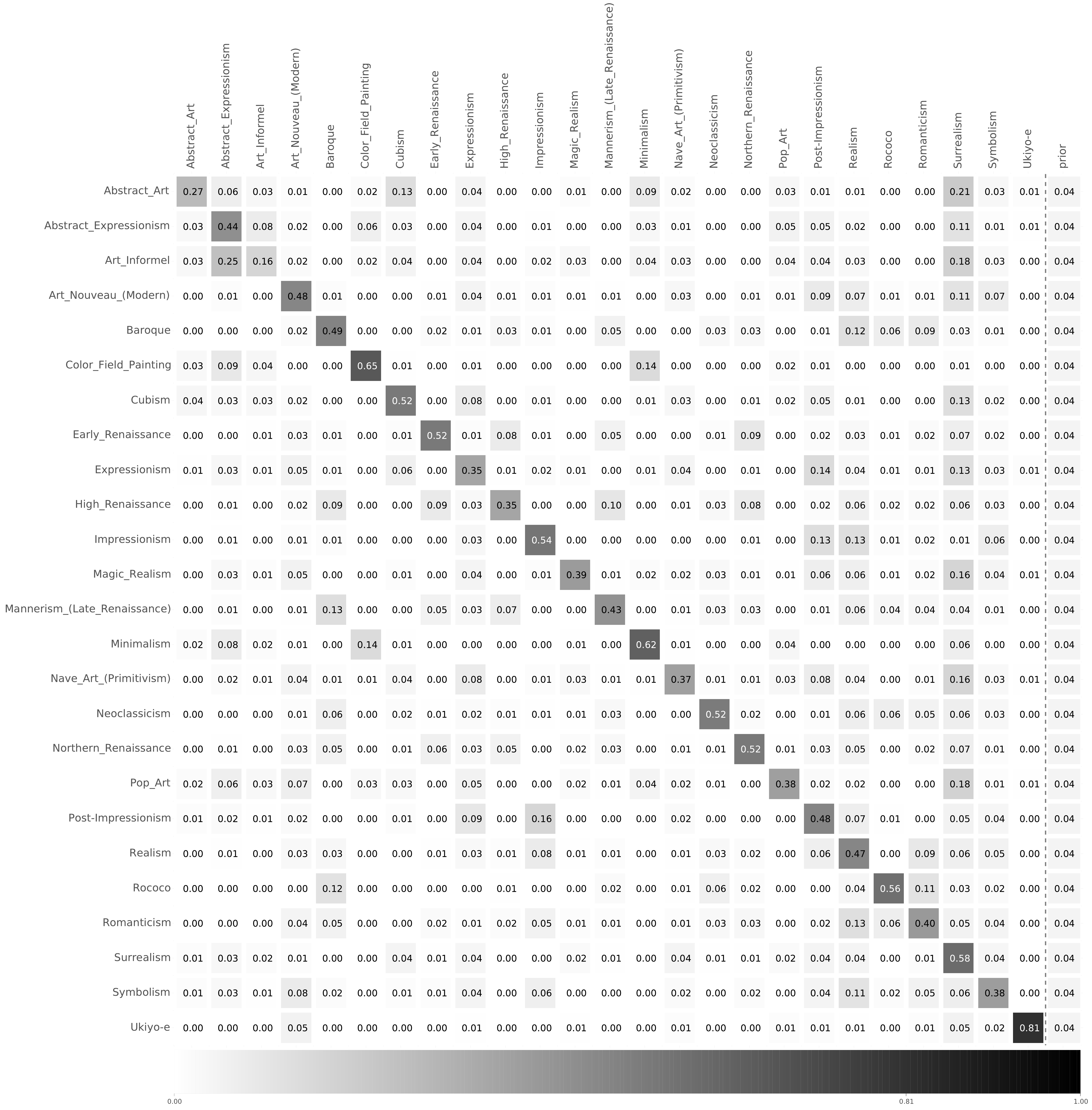}
\caption
[Confusion matrix of our best classifier on the Wikipaintings dataset.]
{Confusion matrix of our best classifier (\mbox{Late-fusion $\times$ Content}) on the Wikipaintings dataset.}
\label{fig:wp_conf}
\end{figure*}

\begin{figure*}[h!]
\centering
\includegraphics[width=\linewidth]{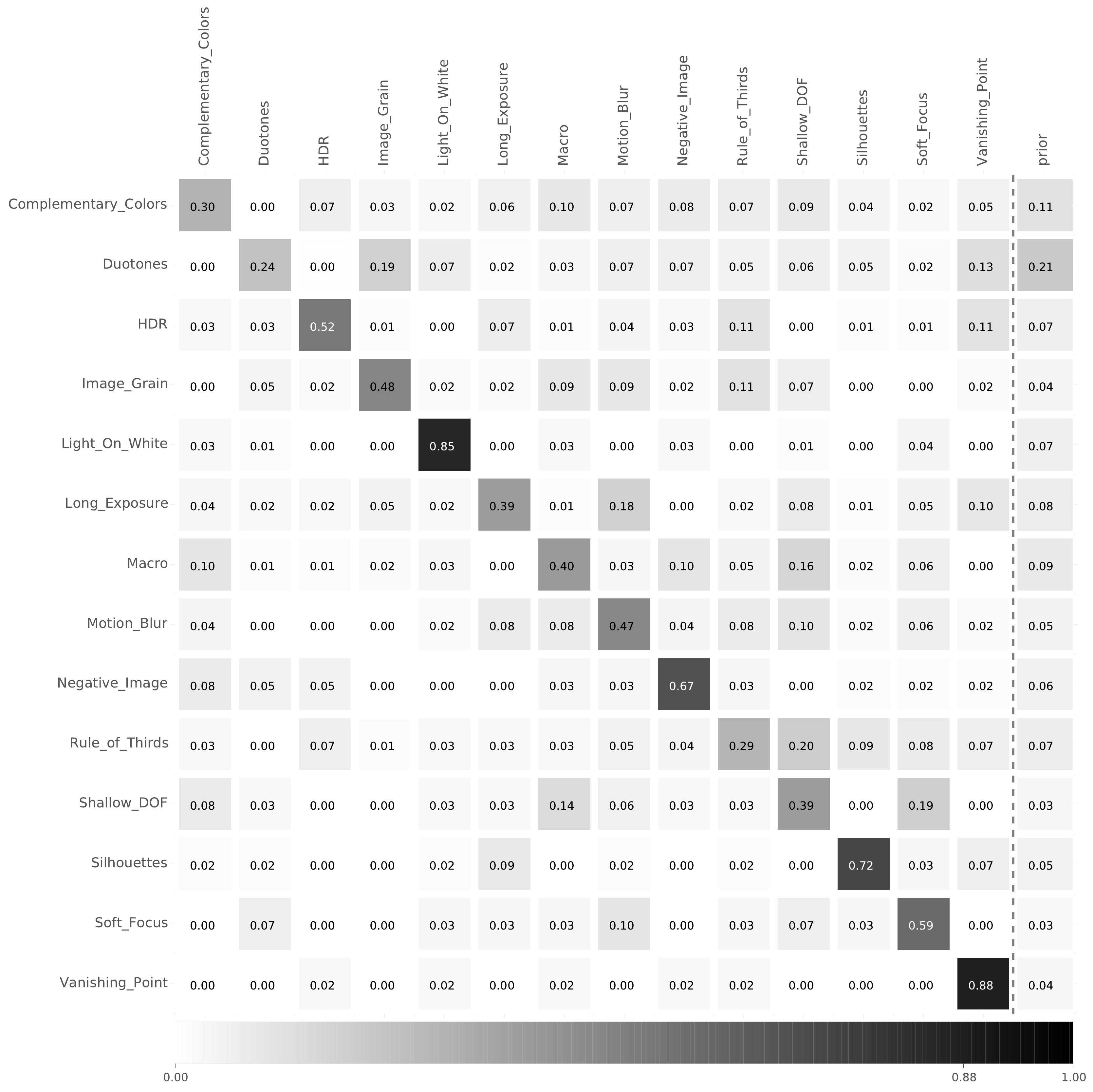}
\caption
[Confusion matrix of our best classifier on the AVA Style dataset.]
{
Confusion matrix of our best classifier (\mbox{Late-fusion $\times$ Content}) on the AVA Style dataset.
The right-most ``prior'' column reflects the distribution of ground-truth labels in the test set.
The confusions are mostly understandable: ``Soft Focus'' vs. ``Depth of Field'' for example.
}
\label{fig:ava_style_conf}
\end{figure*}


\begin{thebibliography}{25}
\providecommand{\natexlab}[1]{#1}
\providecommand{\url}[1]{\texttt{#1}}
\expandafter\ifx\csname urlstyle\endcsname\relax
  \providecommand{\doi}[1]{doi: #1}\else
  \providecommand{\doi}{doi: \begingroup \urlstyle{rm}\Url}\fi

\bibitem[Agarwal et~al.(2012)Agarwal, Chapelle, Dudik, and
  Langford]{Agarwal-JMLR-2012}
Alekh Agarwal, Olivier Chapelle, Miroslav Dudik, and John Langford.
\newblock {A Reliable Effective Terascale Linear Learning System}.
\newblock \emph{Journal of Machine Learning Research}, 2012.

\bibitem[Bergamo and Torresani(2012)]{Bergamo-CVPR-2012}
A.~Bergamo and L.~Torresani.
\newblock {Meta-class Features for Large-Scale Object Categorization on a
  Budget}.
\newblock In \emph{CVPR}, 2012.

\bibitem[Borth et~al.(2013)Borth, Ji, Chen, and Breuel]{Borth-MM-2013}
Damian Borth, Rongrong Ji, Tao Chen, and Thomas~M Breuel.
\newblock {Large-scale Visual Sentiment Ontology and Detectors Using Adjective
  Noun Pairs}.
\newblock In \emph{ACM MM}, 2013.

\bibitem[Datta et~al.(2006)Datta, Joshi, Li, and Wang]{Datta-ECCV-2006}
Ritendra Datta, Dhiraj Joshi, Jia Li, and James~Z Wang.
\newblock {Studying Aesthetics in Photographic Images Using a Computational
  Approach}.
\newblock In \emph{ECCV}, 2006.

\bibitem[Deng et~al.(2009)Deng, Dong, Socher, Li, Li, and
  Fei-Fei]{Deng-CVPR-2009}
Jia Deng, W.~Dong, R.~Socher, L.-J. Li, K.~Li, and Li~Fei-Fei.
\newblock {ImageNet: A Large-Scale Hierarchical Image Database}.
\newblock In \emph{CVPR}, 2009.

\bibitem[Dhar et~al.(2011)Dhar, Ordonez, and Berg]{Dhar-CVPR-2011}
Sagnik Dhar, Vicente Ordonez, and Tamara~L Berg.
\newblock {High Level Describable Attributes for Predicting Aesthetics and
  Interestingness}.
\newblock In \emph{CVPR}, 2011.

\bibitem[Donahue et~al.(2013)Donahue, Jia, Vinyals, Hoffman, Zhang, Tzeng,
  Darrell, Eecs, and Edu]{Donahue2013}
Jeff Donahue, Yangqing Jia, Oriol Vinyals, Judy Hoffman, Ning Zhang, Eric
  Tzeng, Trevor Darrell, Trevor Eecs, and Berkeley Edu.
\newblock {DeCAF: A Deep Convolutional Activation Feature for Generic Visual
  Recognition}.
\newblock Technical report, 2013.
\newblock arXiv:1310.1531v1.

\bibitem[Duchi et~al.(2011)Duchi, Hazan, and Singer]{duchi2011adaptive}
John Duchi, Elad Hazan, and Yoram Singer.
\newblock {Adaptive Subgradient Methods for Online Learning and Stochastic
  Optimization}.
\newblock \emph{Journal of Machine Learning Research}, 2011.

\bibitem[Everingham et~al.(2010)Everingham, {Van Gool}, Williams, Winn, and
  Zisserman]{pascal-voc-2010}
M~Everingham, L~{Van Gool}, C~K~I Williams, J~Winn, and A~Zisserman.
\newblock {The PASCAL VOC Challenge Results}, 2010.

\bibitem[Gygli et~al.(2013)Gygli, Nater, and Gool]{Gygli-ICCV-2013}
Michael Gygli, Fabian Nater, and Luc~Van Gool.
\newblock {The Interestingness of Images}.
\newblock In \emph{ICCV}, 2013.

\bibitem[Harel et~al.(2006)Harel, Koch, and Perona]{Harel-NIPS-2006}
Jonathan Harel, Christof Koch, and Pietro Perona.
\newblock {Graph-Based Visual Saliency}.
\newblock In \emph{NIPS}, 2006.

\bibitem[Isola et~al.(2011)Isola, Xiao, Torralba, and Oliva]{Isola-CVPR-2011}
Phillip Isola, Jianxiong Xiao, Antonio Torralba, and Aude Oliva.
\newblock {What Makes an Image Memorable?}
\newblock In \emph{CVPR}, 2011.

\bibitem[Jia(2013)]{Jia13caffe}
Yangqing Jia.
\newblock {Caffe}: an open source convolutional architecture for fast feature
  embedding.
\newblock \url{http://caffe.berkeleyvision.org/}, 2013.

\bibitem[Joo et~al.(2014)Joo, Li, Steen, and Zhu]{joo2014}
Jungseock Joo, Weixin Li, Francis Steen, and Song-Chun Zhu.
\newblock {Visual Persuasion: Inferring Communicative Intents of Images}.
\newblock In \emph{CVPR}, 2014.

\bibitem[Keren(2002)]{keren2002}
Daniel Keren.
\newblock {Painter Identification Using Local Features and Naive Bayes}.
\newblock In \emph{ICPR}, 2002.

\bibitem[Khosla et~al.(2014)Khosla, Sarma, and Hamid]{khosla2014}
A.~Khosla, A.~Das Sarma, and R.~Hamid.
\newblock {What Makes an Image Popular?}
\newblock In \emph{WWW}, 2014.

\bibitem[Krizhevsky et~al.(2012)Krizhevsky, Sutskever, and
  Hinton]{krizhevsky2012imagenet}
Alex Krizhevsky, Ilya Sutskever, and Geoff~E. Hinton.
\newblock {ImageNet Classification with Deep Convolutional Neural Networks}.
\newblock In \emph{NIPS}, 2012.

\bibitem[Li and Chen(2009)]{Li-SP-2009}
Congcong Li and Tsuhan Chen.
\newblock {Aesthetic Visual Quality Assessment of Paintings}.
\newblock \emph{IEEE Journal of Selected Topics in Signal Processing},
  3\penalty0 (2):\penalty0 236--252, 2009.

\bibitem[Lyu et~al.(2004)Lyu, Rockmore, and Farid]{Lyu-PNAS-2004}
Siwei Lyu, Daniel Rockmore, and Hany Farid.
\newblock {A Digital Technique for Art Authentication}.
\newblock \emph{PNAS}, 101\penalty0 (49), 2004.

\bibitem[Marchesotti and Perronnin(2013)]{Marchesotti-BMVC-2013}
Luca Marchesotti and Florent Perronnin.
\newblock {Learning beautiful (and ugly) attributes}.
\newblock In \emph{BMVC}, 2013.

\bibitem[Mensink and van Gemert(2014)]{Mensink2014}
Thomas Mensink and Jan van Gemert.
\newblock {The Rijksmuseum Challenge: Museum-Centered Visual Recognition}.
\newblock In \emph{ICMR}, 2014.

\bibitem[Murray et~al.(2012)Murray, Barcelona, Marchesotti, and
  Perronnin]{Murray-CVPR-2012}
Naila Murray, De~Barcelona, Luca Marchesotti, and Florent Perronnin.
\newblock {AVA: A Large-Scale Database for Aesthetic Visual Analysis}.
\newblock In \emph{CVPR}, 2012.

\bibitem[Oliva and Torralba(2001)]{Oliva-IJCV-2001}
Aude Oliva and Antonio Torralba.
\newblock {Modeling the Shape of the Scene: A Holistic Representation of the
  Spatial Envelope}.
\newblock \emph{IJCV}, 42\penalty0 (3):\penalty0 145--175, 2001.

\bibitem[Palermo et~al.(2012)Palermo, Hays, and Efros]{Palermo-ECCV-2012}
Frank Palermo, James Hays, and Alexei~A Efros.
\newblock {Dating Historical Color Images}.
\newblock In \emph{ECCV}, 2012.

\bibitem[Shamir et~al.(2010)Shamir, Macura, Orlov, Eckley, and
  Goldberg]{shamir2010}
Lior Shamir, Tomasz Macura, Nikita Orlov, D.~Mark Eckley, and Ilya~G. Goldberg.
\newblock {Impressionism, Expressionism, Surrealism: Automated Recognition of
  Painters and Schools of Art}.
\newblock \emph{ACM Trans.~Applied Perc.}, 7\penalty0 (2), 2010.

\end{thebibliography}
\end{document}